\pdfoutput=1
\documentclass[11pt]{article}

\usepackage{EMNLP2023}

\usepackage{times}
\usepackage{latexsym}

\usepackage[T1]{fontenc}
\usepackage[utf8]{inputenc}

\usepackage{microtype}

\usepackage{inconsolata}

\usepackage{ dsfont }
\usepackage{graphicx}
\usepackage{amsmath}
\usepackage{nccmath}
\usepackage{ amssymb }
\usepackage{float}
\usepackage{multirow}
\usepackage{makecell}

\title{Longer Fixations, More Computation:

Gaze-Guided Recurrent Neural Networks}

\author{Xinting Huang\textsuperscript{*} \quad Jiajing Wan\textsuperscript{*} \quad Ioannis Kritikos \quad Nora Hollenstein \\ University of Copenhagen \\ \texttt{\{wbn969, djk272, pvg847\}@alumni.ku.dk, nora.hollenstein@hum.ku.dk}}
\begin{document}
\maketitle

\def\thefootnote{*}\footnotetext{Equal contribution}\def\thefootnote{\arabic{footnote}}

\begin{abstract}
Humans read texts at a varying pace, while machine learning models treat each token in the same way in terms of a computational process. Therefore, we ask, does it help to make models act more like humans? In this paper, we convert this intuition into a set of novel models with fixation-guided parallel RNNs or layers and conduct various experiments on language modeling and sentiment analysis tasks to test their effectiveness, thus providing empirical validation for this intuition. Our proposed models achieve good performance on the language modeling task, considerably surpassing the baseline model. In addition, we find that, interestingly, the fixation duration predicted by neural networks bears some resemblance to humans' fixation. Without any explicit guidance, the model makes similar choices to humans. We also investigate the reasons for the differences between them, which explain why ``model fixations'' are often more suitable than human fixations, when used to guide language models.
\end{abstract}

\section{Introduction}

Eyes are an integral part of the intelligence of human beings, as they are highly connected with brain mechanisms and cognitive processes, such as attention \citep{sood2020improving}, memory \citep{ryan2020eyes,johnson2014task} and decision-making \citep{spering2022eye}  enabling humans to visually process and interact with the environment. Since eyes are a window to the brain and mind \citep{shaikh2018eye} eye-tracking data, data that assess eye movements which are processed by eye-tracking devices, can be an invaluable information source. In recent years, eye-tracking data has been utilized to facilitate the aims of researchers in a wide range of fields, from the medical field and psychology to the sports and Natural Language Processing (NLP) field \citep{harezlak2018application, mele2012gaze, rahal2019understanding}.

In this paper, we try to make machine learning models imitate human behavior when reading text. That is, human eyes fixate on words for different durations of time, which indicate the workload or the amount of processing for the human brain. We propose novel neural network architectures that are able to adjust the computational process for each word, following the guidance of eye fixation duration. In general, we find that human fixation data show limited effectiveness for the specific model architectures designed by us. However, the varying computational processes and fixation duration mechanism inspired by fixation data produce promising results, when using the ``model fixations''. 

We evaluate our model performance on a language modeling and on a sentiment analysis task. Our best model achieves better performance than the best normal recurrent neural networks, whose test perplexity on WikiText-2 are 70.5 and 76.0, respectively.\footnote{Code available at \\
\texttt{https://github.com/huangxt39/FG-RNN}}

In addition to the performance gains, we also observe that the ``model fixations'' is somewhat similar to human fixations. In other words, neural networks make similar decisions to humans. This finding not only supports our assumption that the proposed architecture can behave meaningfully but also justifies the usage of cognitive data in NLP research.

\section{Related Work}

In recent years, NLP frameworks have increasingly integrated eye-movement attributes, such as gaze points, fixations, saccades, and scanpaths \cite{ijcai2020p0683}. The eye-tracking data benefits NLP models both in performance and interpretability. It has been shown that eye-tracking features can help improve prediction in tasks such as named entity recognition \cite{hollenstein2019entity, tokunaga2017eye}, part-of-speech tagging \cite{barrett-etal-2018-unsupervised}, sentiment analysis \cite{mishra2017learning} as well as general NLP benchmark tasks \cite{khurana2023synthesizing}. On the other hand, eye-tracking data is also  used to investigate the relationship between human behavior and neural attention \cite{hahn2018modeling, sood2020interpreting, brandl-hollenstein-2022-every}. 
 
Recurrent neural networks (RNNs) have shown an impressive capacity for cognitive-behavioral data while dealing with different language processing tasks. \citet{klerke2016improving} leveraged eye-tracking corpora to improve sentence compression models and apply LSTM as the recurrent unit for capturing contextual information with multitask learning. The model successfully learns to perform sentence compression by learning to predict eye gaze, imitating in a way the human reads when the human gaze ignores less important information and assigns higher attention to the relevant content. \citet{hollenstein2019entity} made use of 17 gaze features from three gaze corpora to augment a recurrent neural named entity recognition system with gaze features showing substantial improvements over the baselines.

While Transformer \cite{46201} models are popular nowadays, RNNs still have advantages in some cases. \citet{frank2019interaction} proved that RNN-based models could simultaneously learn about structural and semantic relations between words and show identical patterns to human reading on the same sentences. \citet{gulordava2018colorless} suggested that RNNs are able to acquire deeper grammatical competence while providing more reliable predictions with the long-distance agreement. \citet{baroni2022proper} also found that LSTMs succeed at long-distant agreement with high accuracy. \citet{warstadt2022artificial} suggested that in grammar learning, the differences between the biases of LSTMs and Transformers might be weaker than people expect. Therefore, there is still a lot of space to explore in the RNN structure while approaching human language processing with cognitive-behavioral data.

\section{Model}

In order to make neural networks behave like humans, i.e., with different processing workloads on different words, we design different recurrent architectures that have the flexibility to treat words differently. In general, the common characteristic is that there is more computation on the words associated with longer fixation duration. The specific models used in each task are described in Appendix \ref{ap:perp}.

\subsection{Fixation-Guided Parallel RNN}
The vanilla RNN takes a sequence of tokens as input, and each time one element is fed into the model. It is the same for our fixation-guided parallel (FGP) RNN,  but it calculates the following function:
\begin{equation} 
\medmath{
\label{FG_P}
    h_t^k = \left \{ 
    \begin{array}{ll}
       h^k_{t-1} & k > d_t \\
       \phi(W^{k}_{ih} x_t  + b^k_{ih} + W^{k}_{hh} h^k_{t-1}  + b^k_{hh})   & k \leq d_t
    \end{array}
     \right.}
\end{equation}
The second branch of the function is exactly the same as vanilla RNN, where $h_t $ is the hidden state at time $t$, $x_t$ is the input at time $t$, and $h_{t-1}$ is the hidden state at time $t-1$. $\phi(\cdot)$ is a non-linear activation function applied to each entry of the vector. $W^T_{ih}$ and $b_{ih}$, $W^T_{hh}$ and $b_{hh}$ are the trainable parameters of fully-connected layers applied to the input and the hidden state respectively.

 This model has multiple components which are indexed by $k$, where $k \in \{1, 2, \dots, K\}$. $K$ denotes the number of components of the model.
 Each component works like a normal RNN, so there are $K$ hidden states at each time, and correspondingly multiple groups of trainable parameters. As indicated by Equation (\ref{FG_P}), the model is controlled by the fixation duration $d_t$ associated with the input token. $d_t$ is the eye fixation duration of the input token at time $t$, we use total reading time (TRT) as the duration instead of the number of fixations or any other metrics. To better illustrate our model, we start with the simplest case,  where $d_t$ is an integer. The original fixation duration is mapped into a smaller discrete space, $d_t \in \{1,2, \cdots K\}$, and greater $d_t$ represents a longer fixation duration. 

\begin{figure*}[h]
\includegraphics[width=0.9\textwidth]{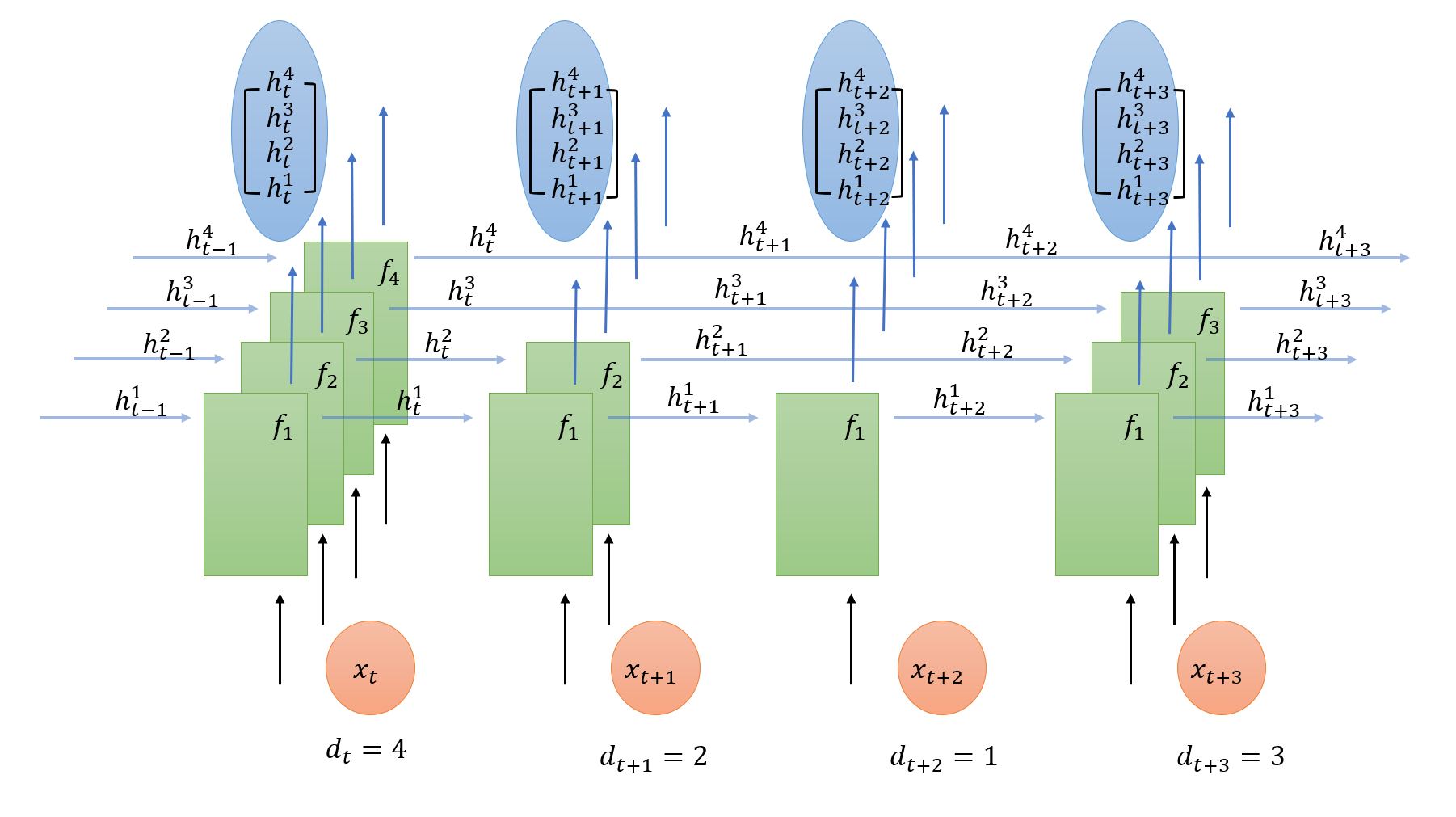}
\caption{Visualization of fixation-guided parallel RNN with the number of components $K=4$. The fixation duration $d_t$ for each token is a discrete and fixed integer. This figure explains the Equation (\ref{FG_P})}
\label{fig:FG_P}

\end{figure*}

Figure \ref{fig:FG_P} shows the case where $K=4$, $f_k$ represents the function of a single RNN. This figure illustrates Equation (\ref{FG_P}), with green squares representing the cases where $k \leq d_t$. When there are no green squares, the hidden state vector is directly passed to the next step, representing the cases where $k > d_t$.

In the fixation-guided parallel RNN, the hidden state at each time $t$ is a collection of hidden state vectors $[h_t^1, h_t^2, \cdots, h_t^K]$. When the model is used together with other neural layers, the output of the model that is fed into the next layer should be the concatenation of all these hidden state vectors. As described in Equation (\ref{FG_P}), the longer the fixation duration is, the more hidden state vectors are updated. When $d_t$ achieves the highest value, all hidden state vectors are updated. Intuitively, when a word occurs with a long fixation duration, it is usually important or contains new information that can not be inferred from the context \cite{kliegl2004length}, it correspondingly updates more of the model's memory. In contrast, the words with short fixation duration are usually of less importance and thus should affect the model to a limited extent.

\subsection{Stacked Fixation-Guided Parallel RNN}
Like normal RNN, there can be a stacked version of proposed FGP RNN. The stacked FGP RNN is as follows:
\begin{align}
 \label{stacked-FGP-rnn}
  h_t^{k(l)} &= \left \{ 
    \begin{array}{ll}
  h^{k(l)}_{t-1} & k > d_t \\
   & \\
       \phi(W^{k(l)}_{ih} x_t^{(l)}  + b^{k(l)}_{ih} + &  \\ W^{k(l)}_{hh} h^{k(l)}_{t-1} + b^{k(l)}_{hh})   &  k \leq d_t
    \end{array}
     \right. \\
     x_t^{(l)} & = W_{oh}^{(l-1)} h_t^{(l-1)} + b_{oh}^{(l-1)}
\end{align}
where $(l)$ is the layer index. $h_t^{(l-1)}$ is the concatenation of all the hidden states from the last layer $h_t^{(l-1)} = [h_t^{1(l-1)}, h_t^{2(l-1)}, \cdots, h_t^{K(l-1)}]$. In our stacked FGP RNN, we apply a fully-connected layer (with parameter $W_{oh}^{(l-1)}$ and $b_{oh}^{(l-1)}$) to the concatenated hidden state to reduce the dimensionality and then feed it forward to the next layer. Note that we use the same number of components $K$ for each layer, and use the same $d_t$ repeatedly at each layer. 

\subsection{Fixation-Guided Parallel LSTM}
Analogously, we can apply the same idea to a Long Short-Term Memory (LSTM) network, since it is also a kind of recurrent neural network. More specifically, in our FGP LSTM, we apply the fixation-guided update only to the cell state. That is:
\begin{align}
\label{FGP-lstm}
    c^k_t & = \left \{ 
    \begin{array}{ll}
       c^k_{t-1} & k > d_t \\
       f^k_t \odot c^k_{t-1} + i^k_t \odot \Tilde{c}^k_t   & k \leq d_t
    \end{array}
     \right. \\
    h^k_t & = o^k_t \odot \mbox{tanh}(c^k_t)
\end{align}
 where $h^k_t$ is the hidden state of component $k$ at time $t$, $c^k_{t-1}$ and $c^k_t$ is the cell state of component $k$ at time $t-1$ and $t$ respectively. $i_t^k, f_t^k, o_t^k, \Tilde{c}^k_t$ are the input, forget, output gates and cell input respectively. $\odot$ is the Hadamard product. It is the same equation as vanilla LSTM when $k \leq d_t$.
% Similar to the case of vanilla RNN, the fixation duration is used to control the update of the cell state. For those components which simply pass the last cell state forward ($k > d_t$), we can see that it is unnecessary to calculate $i_t^k$, $f_t^k$, and $g_t^k$, but the $h_t^k$ is not necessarily equal to $h_{t-1}^k$ in this case, which  is different from the FGP of vanilla RNN. 

% The output of FGP LSTM is also the concatenation of the hidden state of all components, i.e., $h_t = [h_t^{1}, h_t^{2}, \cdots, h_t^{K}]$. For the multi-layer FGP LSTM, we also apply a fully connected layer to $h_t$ and then feed it forward to the next layer.

\subsection{Fixation-Guided-Layer RNN}
In the previous section, we introduce a kind of parallel structure of RNN, and the fixation duration controls the workload or the amount of processing in terms of the number of activated parallel RNNs. In this section, we introduce a similar structure, while the fixation duration controls the processing in terms of the number of activated layers.

Our Fixation-Guided-Layer (FGL) RNN takes the following form:
\begin{equation}
\label{FGL-rnn}
     h_t^{(l)} = \left \{ 
    \begin{array}{ll}
       h_{t-1}^{(l)} & l > d_t \\
       & \\
       \phi (W^{(l)}_{ih} h_t^{(l-1)}   + b_{ih}^{(l)} & \\ + W^{(l)}_{hh} h_{t-1}^{(l)}  + b_{hh}^{(l)})   & l \leq d_t
    \end{array}
     \right.
\end{equation}\\
where $l$ is the layer index ranging from 1 to $L$. In the simplest case, the fixation duration is represented by integers $d_t \in \{1,2, \cdots L\}$. 
Note that the output of the FGL should be all the hidden states from all the layers, namely $h_t = [h_t^{(1)}, h_t^{(2)}, \cdots, h_t^{(L)}]$. The same intuition still holds in this model, the longer the fixation duration for the current token, the more impact it can make on the hidden state of the model, and the more of its information can flow into the memory. 

It is important to distinguish FGL RNN and Stacked FGP RNN. On the one hand, FGL RNN changes the number of activated layers (by activation we mean executing the full process of computation instead of keeping things unchanged) at each time step. On the other hand, Stacked FGP RNN keeps the same number of layers all the time, while changing the number of parallel component RNNs according to fixation duration data.

Analogously, FGL LSTM is defined as follows:
\begin{align}
   \label{FGL-lstm}
    c^{(l)}_t & = \left \{ 
    \begin{array}{ll}
       c^{(l)}_{t-1} & {(l)} > d_t \\
       f^{(l)}_t \odot c^{(l)}_{t-1} + i^{(l)}_t \odot \Tilde{c}^{(l)}_t   & {(l)} \leq d_t
    \end{array}
     \right. \\
     h^{(l)}_t & = \left \{ 
    \begin{array}{ll}
       h^{(l)}_{t-1} & {(l)} > d_t \\
       o^{(l)}_t \odot tanh(c^{(l)}_t)   & {(l)} \leq d_t
    \end{array}
     \right.
\end{align}

\subsection{Fixation Prediction Model}
\label{FP-model}
In order to use the introduced FGP structure, it is necessary to have fixation duration time on the word level, which is not available in most datasets. Thus, we develop a neural network model to predict the fixation duration on any natural language dataset. This model is trained on eye-tracking recordings from subjects reading natural sentences, using corpora described in Section \ref{eye-corpora}. In this regression task, the model is required to predict the fixation duration on each token, given the sequence of tokens as input. 

We use a one-layer LSTM network as our fixed fixation prediction (fixed FP) model across all experiments, with an embedding layer, and two fully-connected layers on top of it. The model is trained with Mean Squared Error (MSE) as a loss function. Note that the training labels here are not the original fixation duration, but rather the ones after preprocessing, which are finite integers. (See Appendix \ref{preprocess})

By utilizing the fixation prediction model, we can generate quasi-fixation duration on any natural language dataset. In this setting, once the fixation  duration is predicted, it is fixed and becomes part of the dataset. Thus, it can be repeatedly used for all training epochs, as well as when fine-tuning hyper-parameters.

\subsection{Adaptive Fixation Prediction Model}
\label{Adapt-FP-model}
In the previous section, the FP model is treated as a separate part of our fixation-guided models (e.g., FGP RNN). In other words, once it is trained it is not updated anymore and is used only in data preprocessing. Nevertheless, it is also interesting to adapt the FP model to the task that the fixation-guided model is performing.

We incorporate the FP model into the fixation-guided models. More specifically, when the fixation-guided model is trained on a specific task, the parameters of the FP model are updated together with the fixation-guided model. The FP model and fixation predictions are flexibly adapted to the task. In order to make this possible, we need to modify our equations to make the process differentiable, thus, the gradients can be backpropagated to the FP model.

We take the FGP RNN as an example and give the modified version of Equation (\ref{FG_P}). We start by normalizing the prediction of FP model to fit it into the range of $[0,K]$ (see Appendix \ref{ap:model}). We then use the normalized fixation duration $\Bar{d_t}$ to obtain coefficient $\alpha_t^k$.
\begin{equation}
\label{sigmoid}
    \alpha_t^k = \mbox{Sigmoid} \left( (k - 1 - \Bar{d_t}) \cdot s \right)
\end{equation}
where $k \in \{1,2, \cdots, K\}$, $s>1$ is a scalar value in order to make the Sigmoid function curve steeper. For example, for a $\Bar{d_t} \in (0,1)$, the corresponding $[\alpha^1_t, \alpha^2_t, \cdots, \alpha^K_t]$ is close to $[0, 1, 1, \cdots, 1]$. For a $\Bar{d_t} \in (1,2)$, the corresponding $[\alpha^1_t, \alpha^2_t, \cdots, \alpha^K_t]$ is close to $[0, 0, 1, \cdots, 1]$ 

Then the coefficients $\alpha_t^k$ are applied as follows:
% \begin{equation}
% \begin{split}
% \label{adapt-FG_P-2}
%     h_t^k = (1-\alpha_t^k) \cdot \phi(W^{k}_{ih} x_t  + b^k_{ih} \\ 
%     + W^{k}_{hh} h^k_{t-1}  + b^k_{hh}) +  \alpha_t^k \cdot h^k_{t-1}
% \end{split}
% \end{equation}

\begin{equation}
    \label{adapt-FG_P-2}
    \begin{array}{ll}
       h_t^k =  &  (1-\alpha_t^k) \cdot \phi(W^{k}_{ih} x_t  + b^k_{ih} \\ 
         & + W^{k}_{hh} h^k_{t-1}  + b^k_{hh}) +  \alpha_t^k \cdot h^k_{t-1}
    \end{array}
\end{equation}

In sum, we keep the process similar to the one described in equation \ref{FG_P}, while making it differentiable for $\hat{d}_t$. In addition, similar equations can be derived for other fixation-guided models. Importantly, the adaptive FP model is randomly initialized and receives gradient only through $\hat{d}_t$. The human fixation data is not used in this case.

For the architecture of the adaptive FP model, in contrast to the fixed FP model (we refer non-adaptive FP model introduced in Section \ref{FP-model} as the fixed FP model), the adaptive FP model follows the architecture of the fixation-guided model which it is combined with. For example, when it is used together with a single-layer FGP RNN, its main architecture is a one-layer RNN, when it is used together with a 3-layer stacked FGP LSTM, its main architecture is a 3-layer LSTM. Therefore, the capability of the adaptive FP model is able to change together with the fixation-guided model. When a larger and more complex fixation-guided model is used to solve a hard task, the adaptive FP model can also increase its capability so that the fixation-guided model is not held back by it. In addition, it shares the embedding layer with the fixation-guided model.

\subsection{Adaptive Fixation Prediction Model with Multi-Task Learning}
\label{sec:multi-task}
With the adaptive FP model, we can further make use of eye-tracking corpora in a multi-task learning manner. When training on an NLP task, we randomly initialize the two models. There are two learning objectives. One is to minimize the loss on the task $L_1$, the other is to minimize the error on fixation prediction $L_2$. Different from the case described in Section \ref{FP-model}, we use a modified MSE loss, which also takes into account the variance (see Appendix \ref{ap:model2})

In multi-task learning, $L_1$ is a function of the parameters of both models, while $L_2$ is only related to the FP model. We optimize the joint loss function:
\begin{equation}
\label{multi-task}
    L = L_1(\theta_m, \theta_{FP} | D_T) + \lambda \cdot L_2(\theta_{FP} | D_{FP})
\end{equation}
where $\theta_m$ is the parameter of fixation-guided model, $\theta_{FP}$ is the parameter of the FP model. $D_T$ is the data of the task, while $D_{FP}$ is the eye-tracking data. $\lambda$ is the coefficient used to combine the two losses.

\section{Data}
\subsection{Eye-tracking Corpora}
\label{eye-corpora}
For the experiments of this paper, we resort to four eye-tracking data resources: the Dundee corpus \citep{kennedy2003dundee}, the GECO corpus \citep{cop2017presenting}, the ZuCo1 corpus \citep{hollenstein2018zuco} and the ZuCo2 corpus \citep{hollenstein2019zuco}. See Appendix \ref{corpora} for more information about eye-tracking corpora, and Appendix \ref{preprocess} for preprocessing methods.

% \begin{table*}[]
%     \centering
%     \begin{tabular}{c|c|c|c|c} \hline
%          & L1 loss & MSE loss &  Acc (exact match) \% & Acc ($\pm 1$) \% \\ \hline
%      Fixed FP model  & 1.51 & 4.40 & 30.6 & 59.4 \\ \hline

%     \end{tabular}
%     \caption{The performance of the fixed FP model on the test set of collected eye-tracking data. }
%     \label{tab:fixed-FP}
% \end{table*}

\subsection{WikiText-2}
 The WikiText language modeling dataset \cite{merity2016pointer} is a collection of over 100 million tokens extracted from Wikipedia. It is a set of good and featured articles reviewed by humans. The dataset is available in two different sizes, WikiText-2 and WikiText-103. We use WikiText-2 in our experiments. The training set contains 600 articles with around 2 million tokens while the valid and test set contains 60 articles each. 
 
 \subsection{Eye-tracking and Sentiment Analysis-II}
 We used the Eye-tracking and Sentiment Analysis-II dataset \cite{Mishra_Kanojia_Bhattacharyya_2016} for sentiment classification. We choose this dataset because it contains 1000 sentences of text and fixation duration recorded in milliseconds from 7 participants. Therefore, we have access to the true human fixation duration and many interesting experiments can be done. The participants were asked to verbally states the sentiment of this sentence before proceeding to the next. We think this procedure lead subjects to do a task-specific reading. Besides fixation duration, each sentence is presented with binary sentiment labels referring to the majority of participants' annotations.

\begin{table*}[]
\centering
\begin{tabular}{l|lll|lll}
\hline
\multirow{2}{*}{Models}      & \multicolumn{3}{c|}{RNN} & \multicolumn{3}{c}{LSTM} \\
                             & 1M     & 4M     & 16M   & 1M     & 4M     & 16M    \\ \hline \hline
    %single layer \\ \hline 
Vanilla                      & 124.3  & 204.0  & 181.7 & 82.4   & 78.9   & 82.1   \\ \hline
FGP+Fixed FPm                & 108.4  & 111.5  & 137.3 & 83.6   & 77.7   & 75.7   \\
FGP+Adapt FPm+Multi-Task     & 116.0  & 110.3  & 145.1 & 81.0   & 73.4   & 73.1   \\
FGP+Adapt FPm                & 128.8  & 121.3  & 153.1 & \textbf{80.4}   & 73.4   & \textbf{70.5}   \\ \hline \hline
 % multi-layer \\ \hline
3-L Vanilla                  & 118.8  & 117.9  & 135.1 & 88.1   & 77.2   & 76.0   \\ \hline
3-L FGP+Fixed FPm            & 112.4  & 103.0  & 119.9 & 88.5   & 77.6   & 75.6   \\
3-L FGP+Adapt FPm+Multi-Task & 102.3  & \textbf{98.6}   & 107.8 & 83.4   & \textbf{73.1}   & 71.3   \\
3-L FGP+Adapt FPm            & \textbf{102.2}  & 103.7  & \textbf{105.0} & 84.2   & 73.6   & 70.6   \\ \hline
FGL+Fixed FPm                & 121.3  & 122.8  & 158.4 & 90.8   & 84.5   & 83.7   \\
FGL+Adapt FPm+Multi-Task     & 116.3  & 117.2  & 149.0 & 88.2   &  80.5  &     80.6   \\
FPL+Adapt FPm                & 116.4  & 116.7  & 130.1 & 87.0   &  79.1  &    79.1 \\ \hline
\end{tabular}
\caption{Model perplexity on test set for WikiText-2. 3-L stands for 3-layer, FPm stands for FP model. 1M, 4M, 16M are the parameter counts, excluding the parameters of the embedding layer. Note that all the configurations are trained using the same hyper-parameters (except hidden dimension), except for the column RNN 16M, where we use a smaller learning rate.}
\label{tab:big-table}
\end{table*}

\section{Experiments}
We present the models used and their results in this section. The performance of the fixed FP model is in Appendix \ref{perform-FP}. In sum, it predicts fixation duration with an acceptable level of accuracy. The implementation details can be found in Appendix \ref{imple}

\subsection{Configuration}
We first introduce the main experiments conducted on the WikiText-2 dataset, which is a language modeling (LM) task.
% In our experiments, we compare five different structures in various configurations: (a) the vanilla RNN and LSTM neural network, of both single-layer and multi-layer; (b)~the FGP RNN  and LSTM with a fixed FP model; (c) the FGP combined with a randomly initialized adaptive FP model, with and without multi-task learning; (d) FGL RNN and LSTM with a fixed FP model; (e) FGL RNN and LSTM with an adaptive FP model, with and without multi-task learning.
When comparing vanilla RNN with our FGP models, setting the hidden dimension to be the same could be unfair. Since our FGP contains many components, the same hidden dimension results in a significantly larger number of model parameters for FGP RNN. Therefore, we decide to control the number of parameters of the models. We experiment with different numbers of model parameters, namely 1 million, 4 million, and 16 million. Note that the parameters of the embedding layer are excluded in the counting (as well as the last FC layer, since they share weights). We adjust the hidden dimension of the models to fit the required number of parameters.

For the sentiment analysis task, we choose FGP models. Besides the similar configurations as LM, we also use the true fixation duration provided in the dataset. Note that empirically we find multi-layer architecture does not help, so all models' RNN or LSTM has only one layer. In multi-task training, different from before, the fixation prediction task uses the eye-tracking data given by the Sentiment Analysis-II dataset.

\subsection{Results of Language Modeling Task}
\label{result-LM}
% \usepackage{multirow}

We present the results of our main task, LM, in Table \ref{tab:big-table}, from which we can observe the following: 

\textbf{a)} FGP models outperform vanilla models. By comparing vanilla and FGP of the same number of parameters and layers, we can easily see that the best configuration of FGP models consistently outperforms the vanilla models to a substantial extent. We think that the power of FGP models mainly derives from two aspects, the parallel structure and the information provided by eye movement. We explore the contribution of these two aspects in Section \ref{eff-fix}.

\textbf{b)} The best FGL models achieve comparable results with vanilla models. This result shows that activating only a part of RNN layers at each time step does not hinder the model's capability to a large extent. In some cases even better performance can be obtained. However, the result also shows that this approach of leveraging fixation duration may not be advantageous.

\textbf{c)} The adaptive FP model is better than the fixed FP model in most cases. We think the main reason is that the fixed FP model approximates real human fixation duration, which may not be best suited to the task. Humans read the text differently when solving different tasks, so it is not appropriate to guide the machine learning models with the fixation pattern of natural reading for all tasks. Moreover, human fixation is not necessarily the optimal process needed by machine learning models. See more discussion in Section \ref{comparing-human}. For the single-layer FGP RNN, the fixed FP model surpasses the adaptive one. Since the adaptive FP model also uses RNN in this case, this could result from the poor ability of the adaptive FP model's neural network. 

\textbf{d)} Multi-task learning does not strongly affect performance. By comparing all the models with and without Multi-task learning, we can see that the model performance is similar in most cases. For the single-layer FGP RNN model, multi-task learning is consistently helpful, which indicates the coefficient $\lambda$ of Equation (\ref{multi-task}) is non-trivial. We think this is because of the poor ability of the adaptive FP model to converge to the optimal point.

\begin{figure*}[]
\centering
\includegraphics[width=\textwidth]{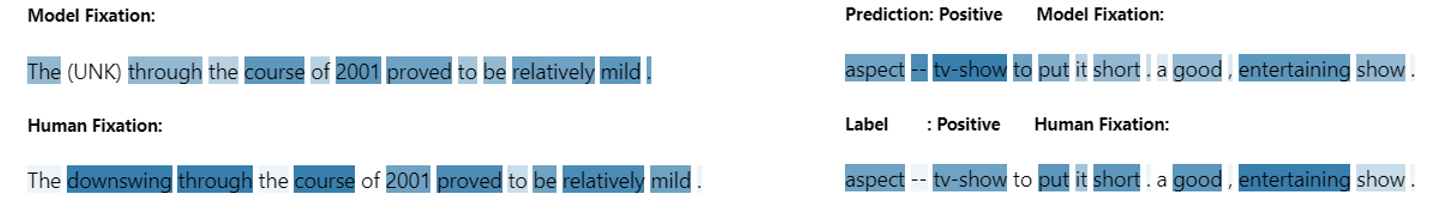}
\caption{Examples of the output of the adaptive FP model (no multitask training), compared with the human fixations. Human fixations are averaged over multiple subjects and are rescaled non-linearly. Deeper color represents longer fixation. The plot on the left side shows the model trained on language modeling dataset. The plot on the right side shows the model trained on sentiment analysis dataset. See more examples in Appendix \ref{ap:figures}.}
\label{fig:adapt-model-fixation}
\end{figure*}

In sum, the best performance is achieved by our single-layer FGP LSTM with Adaptive FP model, with a perplexity of 70.5 on the test set, surpassing the best vanilla model (3-L Vanilla LSTM) by 5.5 (7.2\%). Note that it does not make use of eye-tracking corpora, so the data condition is the same as the vanilla model. A similar result is obtained by its 3-layer counterpart, whose perplexity is 70.6. More importantly, we can see that our best model is very likely to achieve even better performance with more parameters. It seems to have more potential than the best vanilla model.

\subsection{Results of Sentiment Analysis Task}

\begin{table}[]
\centering
\begin{tabular}{lll} \hline
Models                   & \multicolumn{1}{c}{RNN} & \multicolumn{1}{c}{LSTM} \\ \hline \hline
Vanilla                  &   79.9   &   85.2   \\
FGP+Fixed FPm            &  80.3   &    84.3  \\
FGP+TrueFix             &   81.3  &   83.3  \\
FGP+Adapt FPm+Multi-task &  79.8 &  85.9   \\
FGP+Adapt FPm         & 80.1   &  85.5  \\ \hline
\end{tabular}
\caption{Test accuracy of different models for sentiment analysis. Each result is the mean of three runs. ``TrueFix'' denotes the real human fixation data for each sentence.}
\label{tab:SA}
\end{table}

In the Sentiment Analysis-II dataset, the most interesting feature is that real human eye-movement data is available for each sentence. Thus, we can see how the fixation-guided model performs when guided by the true human fixation duration. The results from different models are presented in Table \ref{tab:SA}. We could view a trend that is different from the case of LM. In this case, no model shows significantly better performance than the others. We think there are two important reasons. Firstly, this may be because the average length of sentences is quite short. There is not so much long dependency in the text. Our FGP models' advantage of capturing longer dependency does not manifest itself. Secondly, the dataset only contains 1k sentences. The scarcity of training data could form the bottleneck and strongly limit the models' performance, and make it harder to demonstrate and differentiate the performance of various models.

However, the experiments about this dataset are worth presenting, because it is interesting to see what the model learns from this dataset. Figure \ref{fig:adapt-model-fixation} and \ref{fig:adapt-SAdata} show some examples of ``model fixations'' generated by the adaptive FP model after training the FGP + Adapt FPm on the sentiment analysis dataset. We can see that, without any guidance, the model learns to ``fixate'' on words such as ``entertaining'',  ``horrible'', etc., which are strong indicators of sentiment. In the following section we present our analysis of them.

\subsection{Comparing to Human Fixation Data}
\label{comparing-human}
By examining the fixation durations generated by the adaptive FP model with no multitask training (Figure \ref{fig:adapt-model-fixation}, \ref{fig:adapt-SAdata}, \ref{fig:adapt-ETdata}), we can discern some insights into the difference to human fixation data and the reasons why human fixations are not as helpful as expected. Note that we examine the prediction from the adaptive FP model, which is randomly initialized and learns solely from the task that the fixation-guided model is performing. It is different from the fixed FP model which is trained on human fixation data.

Firstly, even though there is no guidance or hint to lead the model to predict human style fixations, ``model fixations'' bear some resemblance to humans, which we think is an interesting finding. For example, in Figure \ref{fig:adapt-model-fixation}, the fixations on the text ``the course of 2001" in the left plot and ``a good, entertaining" in the right plot. It shows that the neural network model chooses to activate more components for words that require greater cognitive effort from humans. Note that the exact match is impossible since humans themselves read differently from each other \cite{KIDD2018154}. Secondly, as we can see in Figure \ref{fig:adapt-model-fixation} and \ref{fig:adapt-ETdata}, the ``model fixations'' have a smaller variance than human fixations. We think the reason lies mainly in the discrepancy in the tasks that are performed. The model is doing language modeling and predicting each token, so every token is at least important to its next token. On the other hand, humans are trying to understand the sentences during natural reading, so they do not usually fixate on tokens that do not contribute to the sentence comprehension \cite{hahn-keller-2016-modeling}. This also shows a difference between language modeling tasks and the task that humans are performing during reading (more similar to language understanding). Thirdly, low-frequency words require more cognitive workload, so they usually have longer fixation duration \cite{comparingNaming-Schilling, rayner1998eye}. However, in neural models, low-frequency words are likely to be encoded as unknown tokens, which carry almost no information. So it is reasonable that the model chooses to ignore those words. Therefore, it is understandable that using human fixation data does not show much benefit.

\subsection{Effectiveness of Fixation Data}
\label{eff-fix}
In Section \ref{result-LM}, we see a relatively strong performance of our FGP models. We believe this can be attributed to two primary factors: their parallel structure and the use of fixation duration (both human \textit{and} model fixations). To further explore the contribution of these two factors, we conduct experiments with artificial fixation data on the LM task, using the best model in Table \ref{tab:big-table}, i.e., 16M single-layer FGP LSTM. The fixation duration generated by the FP model is replaced by four kinds of artificial values: 1) Random numbers ("Random") in $\{1, 2,\cdots, 12\}$ uniformly distributed, as we use 12 components. They are sampled at the beginning and used repeatedly in each epoch. 2) Random numbers bound to each token type ("Random BT"): one random number is assigned to all occurrences of the same token. 3) Full fixation duration ("Full"): Every token has a maximum fixation duration (i.e., 12). 4) Frequency-based fixation duration ("Freq"): Fixation duration is assigned based on the frequency of each token in the eye-tracking corpora. The higher the frequency, the lower the fixation duration. Meanwhile, we keep the fixation duration uniformly distributed. 

\begin{table}[]
\centering
\begin{tabular}{ll} \hline
Fixation Value & \multicolumn{1}{c}{Perplexity} \\ \hline
Fixed FPm      & 75.7                                \\
Adapt FPm      & 70.5                                \\ \hline
Random         & 77.3                                \\
Random BT      & 76.3                                  \\
Full           & 72.9                                \\
Freq           & 74.7       \\ \hline                         
\end{tabular}
\caption{The perplexity of the single-layer 16M FGP LSTM on the test set of WikiText-2, using different kinds of values as fixation duration.}
\label{fix-exp}
\end{table}

The results are shown in Table \ref{fix-exp}. We can see that replacing fixation duration with random numbers does hinder the model's performance. So good performance does not result from the varying number of activated components at each time step, but from that number varying in an appropriate pattern. More importantly, we observe that with full fixation duration, i.e., all components of the model are working the entire time, the model achieves good results, surpassing the fixed FP model, which indicates the effectiveness of the parallel structure. So the good performance of the FGP model with the adaptive FP model attributes to a considerable extent to the parallel structure, while the adaptive FP model's control contributes to the other important part. The frequency-based fixation also shows a low perplexity, which indicates using frequency is also a good strategy. The results of Full and Freq largely nullify the importance of the fixed FP model. Therefore, the quasi-human-fixation is not as helpful as expected, which can be partly explained by the reasons given in the previous sections. On the other hand, the ``model fixations" generated by the adaptive FP model without the influence of human fixations produces improvements for our FGP models on the LM task.

\section{Conclusion}
In this paper, we explore the idea of making models imitate human behavior when processing text information. We propose a range of novel neural networks, namely Fixation-Guided Parallel and Fixation-Guided-Layer RNNs which have the flexibility to change the computational process according to the fixation duration. We use them together with different fixation values that are generated by fixed or adaptive Fixation Prediction models, and conduct many experiments with different model sizes. We find that better results are achieved by our proposed models. It is also interesting to observe the decision made by the model when it is given free rein, i.e., ``model fixations''. We find that, on one hand, it is similar to human fixations, which strengthens the connection between cognitive science and NLP research. On the other hand, in future research, differences between generative language modeling tasks and human language comprehension should be taken into account when leveraging cognitive signals for NLP.

\bibliography{emnlp2023}
\bibliographystyle{acl_natbib}

\newpage
\appendix

\section{Limitations}
In this paper, we evaluate models on two relatively small datasets due to our limited compute budget. While we believe results on bigger datasets of the same type are similar, it is possible that the performance will become better or worse. Secondly, the training of our adaptive FP model is based on the differentiable controlling of all the RNN components. In Equation (\ref{sigmoid}), the Sigmoid function may not be the best choice, because the gradient from most components is nearly zero (when $\alpha_t^k$ is close to 0 or 1). The model could be further improved by designing a better computation process.

\section{Eye-Tracking Corpora}
\label{corpora}

In general, we take the English part of all the corpora, and only the natural reading data (instead of task-specific reading). All corpora are recorded by professional researchers and devices, with equal or more than 10 subjects reading the text. The text contains multiple domains including news, novels, movie reviews, and Wikipedia articles.

\noindent\newline\textbf{Dundee}
The Dundee Corpus \citep{kennedy2003dundee}
 is an eye-tracking corpus of natural reading of 10 native French and 10 native English participants, with the  English section comprising 58,598 tokens in 2,367 sentences \citep{hollenstein2019entity}.
The subjects read 20 news articles from the newspaper \textit{The Independent} written in their native language. For this paper, we use a pre-processed version of the English version of the Dundee Corpus \citep{barrett2015dundee}.

\noindent\newline\textbf{GECO}
 The Ghent Eye-Tracking Corpus \citep{cop2017presenting} is composed through the reading of the complete Agatha Christie’s novel “The Mysterious Affair at Styles” on behalf of 33 subjects, 14 monolingual British English native speakers and 19 bilingual speakers of Dutch and English. The novel contains 68,606 tokens in 5,424 sentences and the bilingual speakers read half of the novel in their mother tongue and the remaining half in their second language. 

\noindent\newline\textbf{ZuCo1}
The Zurich Cognitive Language Processing Corpus (or ZuCo1 or ZuCo 1.0) \citep{hollenstein2018zuco} is an EEG and eye-tracking corpus of 12 native English speakers reading sentences from 400 movie reviews from the Stanford Sentiment Treebank \citep{socher2013recursive} and 707 biographical sentences  from the Wikipedia relation extraction dataset\citep{culotta2006integrating}. It encompasses EEG and eye-tracking data of 21,629 words in 1107 sentences and 154,173 fixations. 

\noindent\newline\textbf{ZuCo2} The Zurich Cognitive Language Processing Corpus 2.0 (or ZuCo2 or ZuCo 2.0) \citep{hollenstein2019zuco} is an EEG and eye-tracking corpus of 18 native English speakers reading 739 sentences, 349 in a normal reading paradigm and 390 in a task-specific paradigm, all of them extracted from the Wikipedia corpus provided by \citep{culotta2006integrating}.

 \section{Preprocessing}
 \label{preprocess} 
 \subsection{Preprocessing of Eye-Tracking Data}
ZuCo corpus provides eye fixation duration defined by different activities, e.g., first fixation duration (FFD), gaze duration (GD), total reading time (TRT), etc. Other corpora provide similar metrics. We choose the total reading time (TRT) as the fixation duration in this paper, which is available in all corpora introduced before. We use the English data in Dundee, the English data recorded from native English speakers in GECO, the data recorded from task 1 and 2 in ZuCo1, and the data recorded from task 1 in ZuCo2. The chosen data in ZuCo1 and ZuCo2 is the data recorded from natural reading, while in other tasks subjects do the task-specific reading. 

Since the fixation duration distributes differently in different corpora, we normalize the fixation duration for each corpus, by dividing it by the mean duration of the corpus. Moreover, as mentioned in Section \ref{FP-model}, when training the fixed FP model beforehand, we do two steps of processing: 1) take the average over all the subjects and get the mean fixation duration. 2) map the duration values to discrete space $[1,2,\cdots, K]$. In the second step, we take the K-quantiles, that is, we partition the fixation values into K subsets of nearly equal sizes. Each value is mapped to the index of the subset to which it belongs. We think the second step is appropriate since otherwise the FP model learns most for the pattern of the large values. This also avoids the scarcity of the high duration data. When training the FP model in a multi-task setting, we calculate the mean and variance of the fixation duration and use them as described in Equation (\ref{var-loss}).

Note that tokenization plays an important role since the tokenizing method can be different in eye-tracking corpora and NLP tasks. For example, ZuCo1 is tokenized by whitespace. So ``hello!" is considered as one token, and there is one fixation duration value for this token. It's the same for ``I'm". In many NLP tasks, it is common to tokenize ``hello!" into ``hello" and ``!", and tokenize ``I'm" into ``I" and ``'m". It is important to assign the fixation duration in a proper way when tokenizing differently. We follow the following approach: We use a tokenizer from NLTK \cite{nltk} to further tokenize the tokens in eye-tracking corpora, producing one or more new tokens associated with one fixation duration value. If the new token contains any word character, the fixation duration value is assigned to it, otherwise (for punctuation) a small value is assigned ( in our case, 1). In the case where variance is used, the variance of fixation duration is assigned if the new token contains word characters, otherwise, infinite variance is assigned, which allows the FP model to predict any value since we don't know how long the fixation should be. In sum, the word-level fixation duration is converted into token-level.

 \subsection{Preprocessing of Sentiment Analysis-II}
The Sentiment Analysis-II dataset contains two parts of information: the labeling annotation and the fixation duration for the words presented in 
the order they were fixated by each participant. Firstly, we rearranged the  order of the words and filled absent words with 0ms fixation duration according to the original sentence. We, then, used the averaged results from 7 participants for fixation duration. At last, we combined the labeling annotation and the fixation duration by sharing the same sentence text ID. The dataset will include after the preprocessing process the text ID, the word ID in each sequence, the words, the averaged fixation duration for each word, and the classification label.
 
\section{Supplementary Description of Models}

\subsection{Content about Section \ref{Adapt-FP-model}}
\label{ap:model}
In Section \ref{Adapt-FP-model}, we mention that we normalize the predicted values of the adaptive FP model to fit them into the range $[0, K]$, with the following process:
\begin{equation}
\label{adapt-FG_P-1}
    \Bar{d_t} = (\dfrac{\hat{d}_t - \mbox{E}[\hat{d}_t]}{\mbox{std}[\hat{d}_t]} + 1.96) / 3.92 \cdot K
\end{equation}
where $\hat{d}_t$ refers to the predicted fixation duration given by the FP model. $\mbox{E}[\hat{d}_t]$ and $\mbox{std}[\hat{d}_t]$ are the estimated expectation and standard deviation of $\hat{d}_t$. $K$ is the number of component RNN. After subtracting by the mean and dividing by the standard deviation, we assume the values are subjected to the standard normal distribution $\mathcal{N}(0,1)$. Thus, we assume that the $95\%$ of values are distributed in $[-1.96, +1.96]$. By the operations in Equation (\ref{adapt-FG_P-1}), we make most of the normalized duration distributed in $[0,K]$.

Note that the normalization operation here is important. In some preliminary experiments, we observe that when using adaptive FP model without this normalization, the FP model tends to produce similar fixation for all tokens, i.e., small variance. The values are normally concentrated around $K/2$. Thus by applying normalization, we manually enlarge the variance, but the model is still free to choose which token should activate more components. In addition, in the trained adaptive FP model's prediction, the model generates small $\hat{d}_t$ thus resulting in minus $\Bar{d_t}$ for unknown tokens. So the variance in remaining tokens is slightly smaller than it should be after normalization.

\subsection{Content about Section \ref{sec:multi-task}}
\label{ap:model2}
In multi-task training, we do not map the fixation duration to discrete values and take the variance of fixation duration into consideration. The loss of fixation prediction in multi-task learning is defined as:
\begin{equation}
\label{var-loss}
    L_2 = \sum_n \sum_t \frac{(\mbox{E}[d_{t(n)}] - \hat{d}_{t(n)})^2 }{\mbox{Var}[d_{t(n)}]+\epsilon}
\end{equation}
where $\epsilon$ is a small real number to avoid dividing by zero.  $\hat{d}_{t(n)}$ refers to the predicted fixation duration given by the FP model. In contrast to Section \ref{FP-model}, the training label $d_{t(n)}$ here is the original fixation duration. We estimate the expectation $\mbox{E}[d_{t(n)}]$ and variance $\mbox{Var}[d_{t(n)}]$ based on the samples given by multiple subjects. By making use of the variance, we consider the uncertainty of eye movements. If subjects disagree with each other on a word, a prediction different from the mean is more acceptable. Thus, the model learns more about certain fixation data. This loss also avoids the problem that the FP model learns most of the pattern of the long fixations since large values are usually accompanied by large variance.

\subsection{Models for Specific Tasks}
\label{ap:perp}
We apply the aforementioned models to two tasks, one language modeling (LM) task and one sentiment analysis task. In the LM task, the model is to predict the next token based on the previous ones. In other words, the model is to maximize the probability of the given sequence:
\begin{equation}
    p(x_1, \cdots x_n| \theta) = \prod_{i=1}^n p(x_i|x_{<i}, \theta)
\end{equation}

When using our models in the LM task, we add a word embedding layer before the RNN layer, as well as a FC-tanh-FC layer (by FC we mean a fully-connected layer) on the top which generates the final prediction. The second FC layer shares weights with the embedding layer. The performance of the model is measured by perplexity, which is defined as follows:
\begin{equation}
    \mbox{Perplexity} = -\frac{1}{n} \sum_{i=1}^n \log p(x_i|x_{<i}, \theta)
\end{equation}

As for the sentiment analysis task, the model is to classify the given sequence into two categories. The whole model uses the same architecture as in LM, but generates a vector containing probability for both positive and negative labels.

\section{Performance of Fixed Fixation Prediction Model}
\label{perform-FP}

It is important to know whether the fixed FP model is predicting accurate fixation duration. So we report the performance of the fixed FP model on the fixation prediction task. We experiment with many different hyper-parameters and choose the best ones as with our final fixed FP model. The fixed FP model is trained on 75\% of eye-tracking data and tested on the other 25\%. Note that the original fixation duration is mapped into a discrete space $[1,2,\cdots, 12]$. The model is trained with the mapped values. The final L1 loss (mean of the absolute differences between the predicted duration and true duration) on the test set is 1.51, and the MSE loss is 4.40. Figure \ref{fig:FP-example} in the appendix shows an example of the model prediction.

Even though the performance is not excellent, it is acceptable, given the limited amount of eye-tracking data. Most of its predictions are close to the true value, and the mean difference is tolerable. We can assume that the main pattern of eye movement is successfully retained in the prediction. Note that we also find the vocabulary of the model is small, and a non-trivial proportion of tokens in the LM task is encoded as unknown, which may cause worse performance in practice.

\section{Implementation Details}
\label{imple}
We list the hyper-parameters in order of importance. The number of components of FGP models $K$ is 12 in all the related experiments for the LM task. We choose 3 as the number of layers in all multi-layer configurations, following the architecture proposed by \cite{RegLSTM}, including the FGL models. For the adaptive FP models combined with FGP models, the FP models have the same hidden dimension as the FGP components. As for the adaptive FP models combined with FGL models, the hidden dimension of FP models is $1/3$ of that of FGL models. The coefficient $\lambda$ for multi-task learning in Equation (\ref{multi-task}) is set to be 0.3 for all experiments, selected based on empirical experiments. 
The $\epsilon$ in Equation (\ref{var-loss}) is 0.1. The $s$ in Equation (\ref{sigmoid}) is 4.0.

The fixed FP model is a one-layer uni-directional LSTM with the hidden dimension of 100. Its embedding layer is initialized by Glove \cite{pennington2014glove}. It is optimized with Adam optimizer \cite{kingma2014adam} with a fixed learning rate of 0.001. These are the best hyper-parameters we can find for a fixed FP model.

To train and test the fixed FP models, we shuffle and split the collected eye-tracking data into training and test sets, which compose 75\% and 25\% of the total number of examples, respectively. The fixed FP model is trained solely on the training split. Its performance on the test set is described in Section \ref{perform-FP}. However, in multi-task learning, we simply use all the eye-tracking data to train the adaptive FP model. In this case, the average fixation duration is normalized to 0 mean and unit variance, and the variance of fixation duration is normalized accordingly.

We use weight tying \cite{inan2016tying} for all the models used in the LM task. Weight tying shares the weights between the embedding and the softmax layer, reducing the total number of parameters to a large extent. For vanilla, FGP, FGL models, we initialized the word embedding with pre-trained Glove. The dropout rate for all models is 0.5 for the embedding layer and 0.25 for other layers. All models are trained with Adam optimizer using an initial learning rate of 0.001, except for 16M RNN models, in which the learning rate is set to be 0.0001. We use random sequence length introduced by \cite{RegLSTM}, whose mean is 100. The batch size is 64 and the number of training epochs is 50. 

For the sentiment analysis task, we do not control the model parameters to be a certain number, since a larger size does not help in this small dataset. We grid-search over the hyper-parameter space, which is hidden dimension of $[30, 50, 100, 200]$ and component number of $[3, 6, 12]$ and find the optimal combination for each model. We keep other hyper-parameters the same for all sentiment analysis experiments. We use Adam optimizer with a fixed learning rate of 0.001, batch size of 32, and train the models for 30 epochs. The dropout rate is 0.5 for the embedding layer and 0.25 for other layers. For multi-task learning, we use $\lambda$ of 0.001. We shuffle and split the dataset into training and test set, with 20\% of the instances used for testing. Due to the small size of the dataset, the results of the same experiment vary to a non-trivial extent. So we run each experiment three times. The final accuracy is obtained by taking the highest accuracy on the test set over the 30 epochs and then taking the average over three runs.

\counterwithin{figure}{section}

\section{Figures}
\label{ap:figures}
\begin{figure*}[]
\centering
\includegraphics[width=0.7\textwidth]{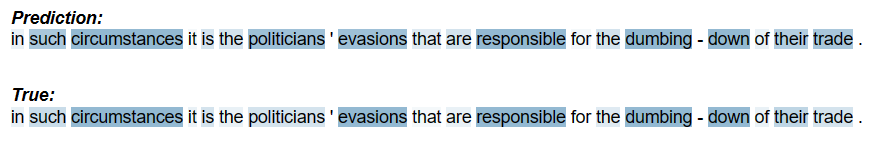}
\caption{An example of the fixation duration predicted by the fixed FP model, compared with the true duration. The model is trained on human fixation data. Text and true fixation duration are from the eye-tracking corpora. Deeper color represents longer fixation}
\label{fig:FP-example}
\end{figure*}

\begin{figure*}[]
\centering
\includegraphics[width=0.95\textwidth]{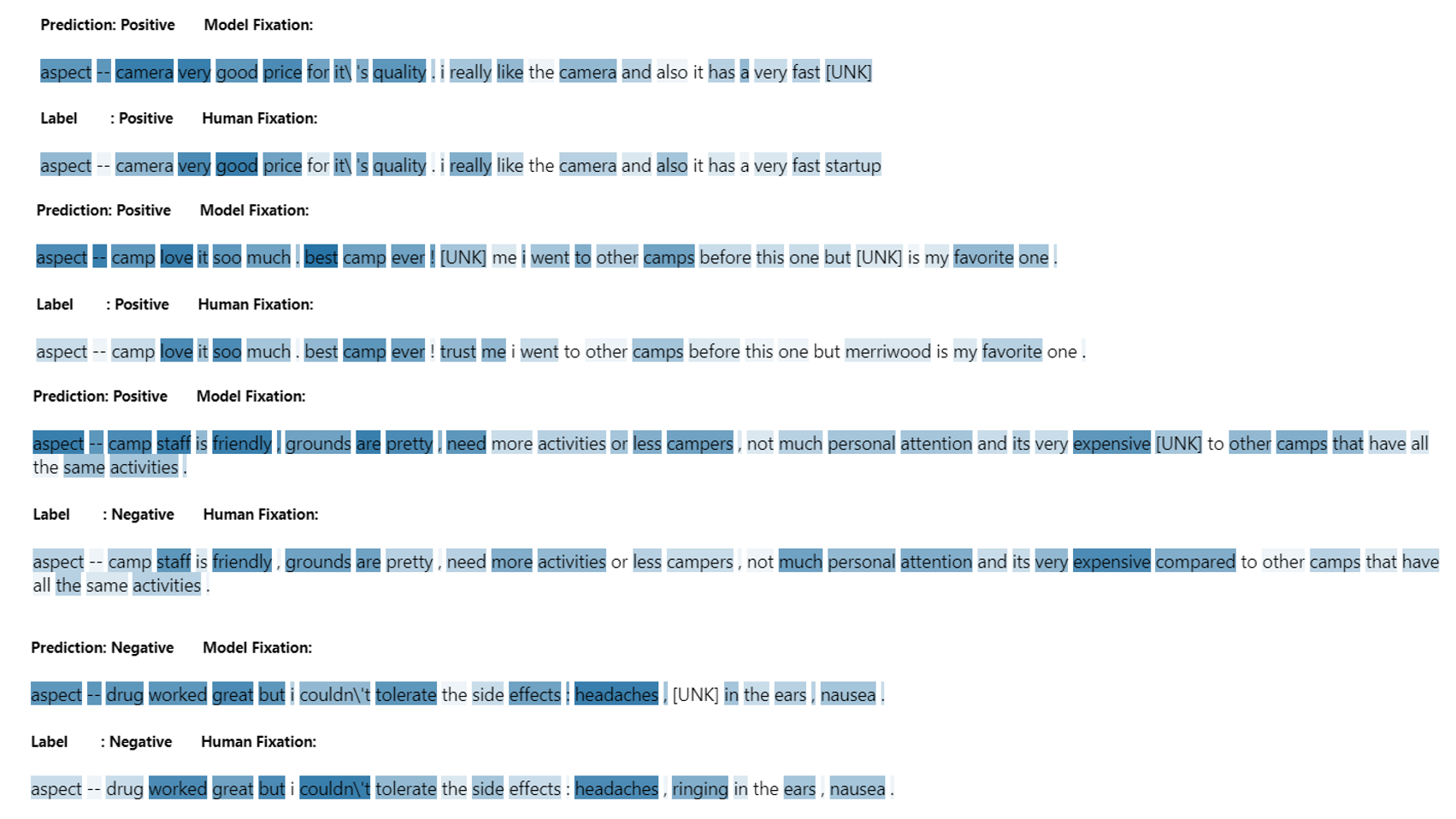}
\includegraphics[width=0.95\textwidth]{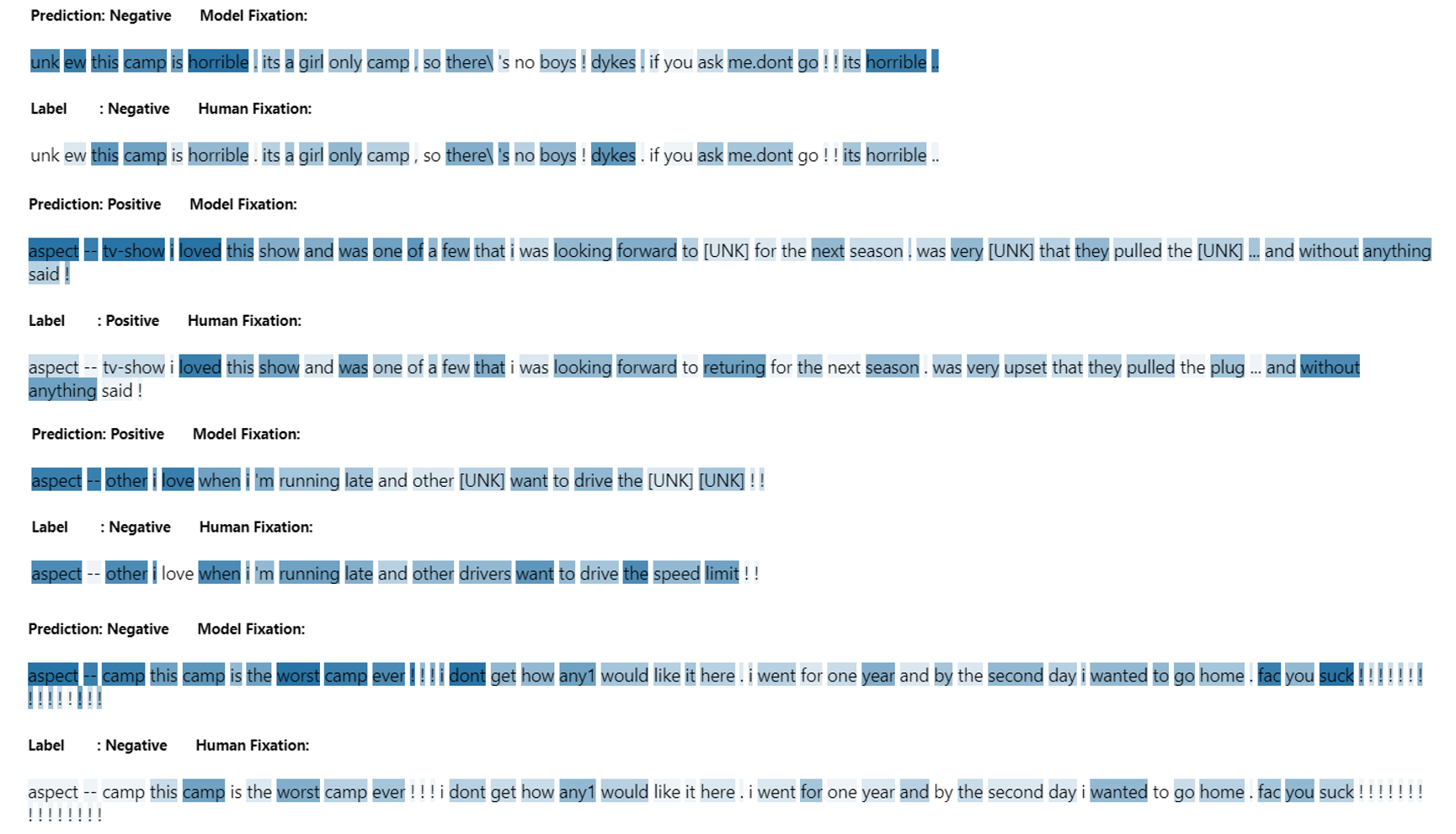}
\caption{Examples of the fixation duration given by the adaptive FP model (no multitask learning), compared with the human fixations. The label and the prediction of the fixation-guided model are also shown. The adaptive FP model is randomly initialized and learns solely from the sentiment analysis dataset. The model is used together with a one-layer FGP LSTM with 3 components and is trained on the Eye-tracking and Sentiment Analysis-II dataset. Input text and true fixation duration shown in the examples are also from the dataset. Note that because the dataset is relatively small, the vocabulary derived from the training set is quite small (2.5k) and does not cover many common words.}
\label{fig:adapt-SAdata}
\end{figure*}

\begin{figure*}[]
\centering
\includegraphics[width=\textwidth]{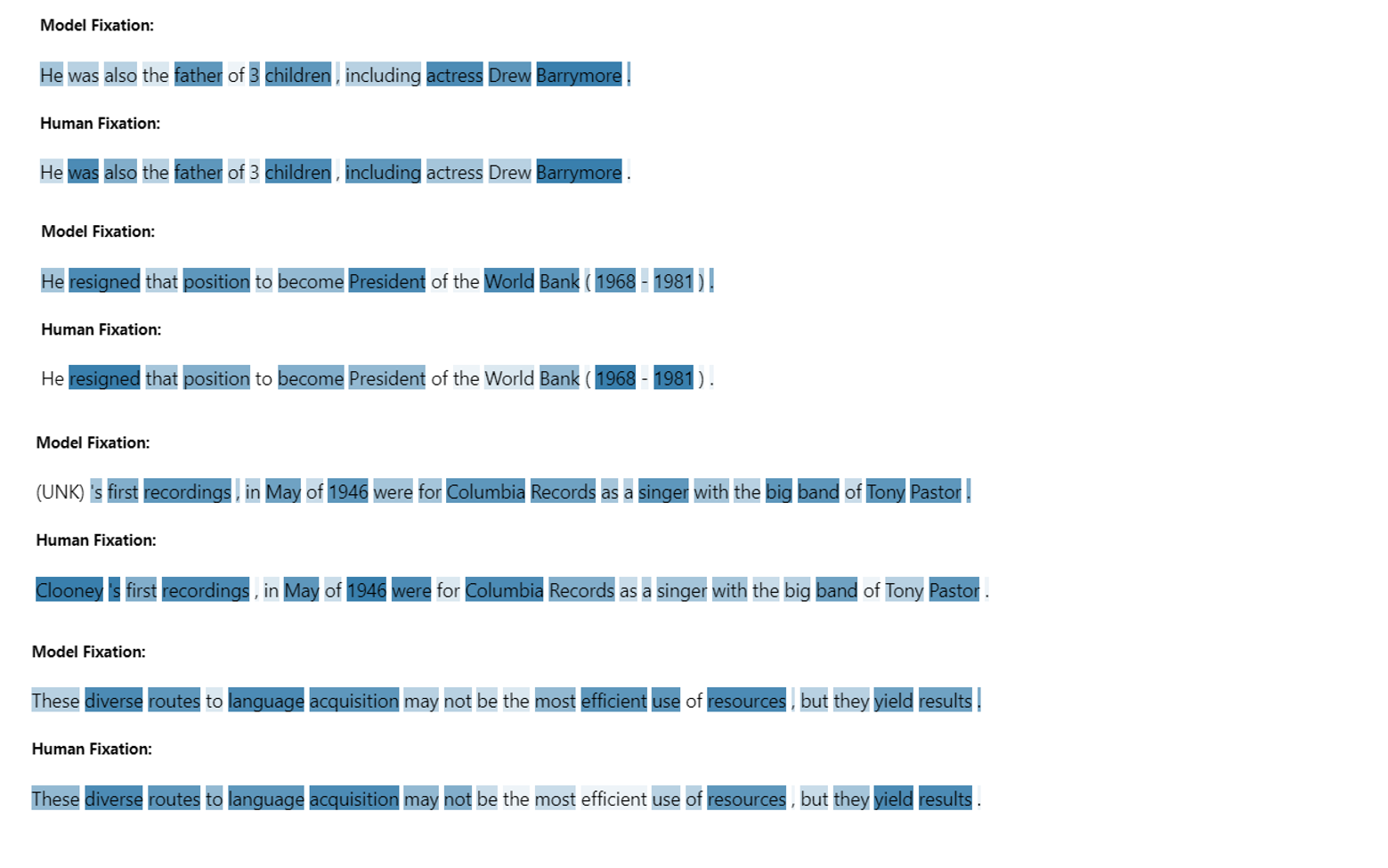}
\includegraphics[width=\textwidth]{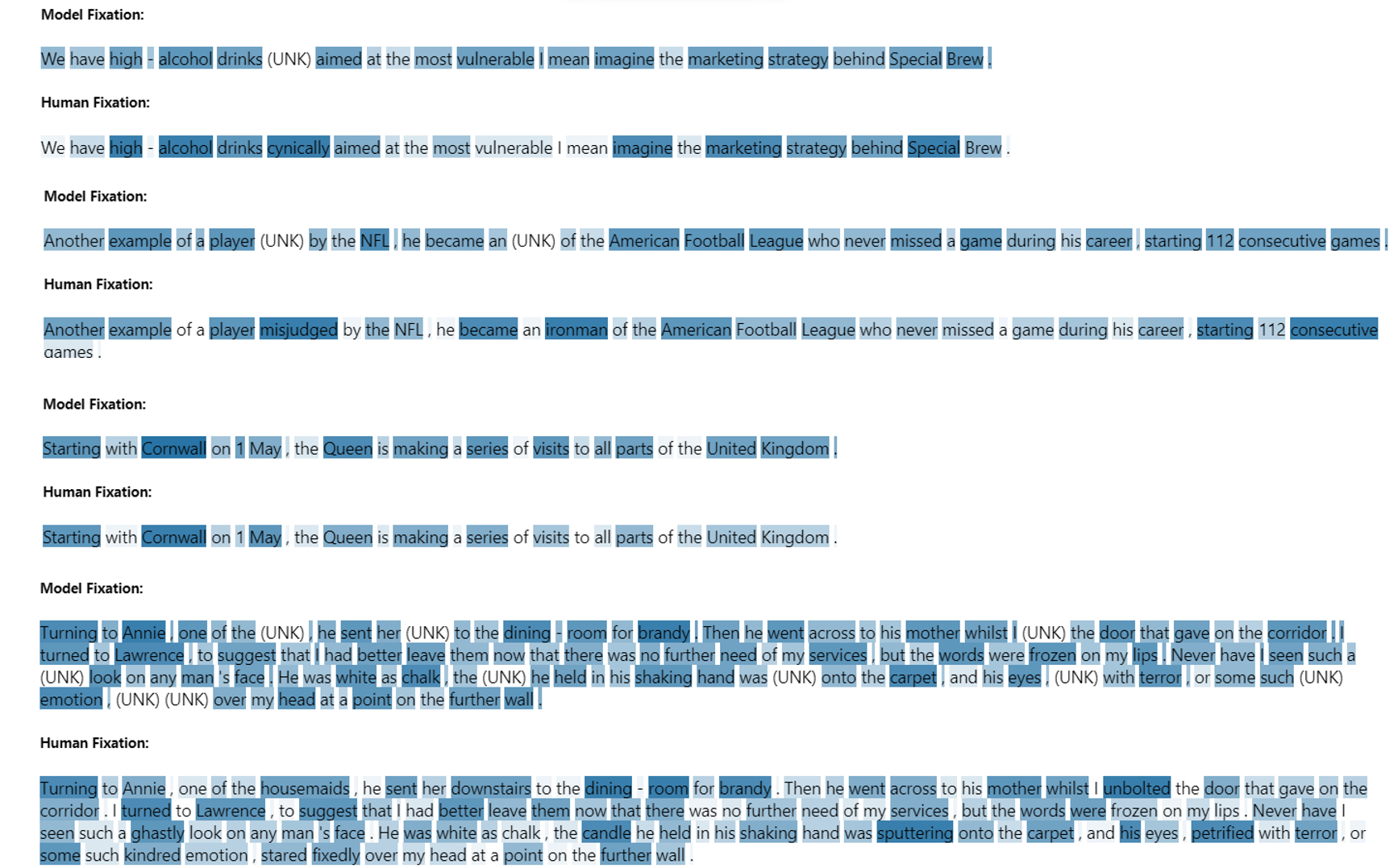}
\caption{Examples of the fixation duration given by the adaptive FP model (no multitask learning), compared with the human fixations. The adaptive FP model is randomly initialized and learns solely from the language modeling dataset. The model is used together with a 16M one-layer FGP LSTM with 12 components and is trained on WikiText-2. But the input text and true fixation duration shown here are from the eye-tracking corpora.}
\label{fig:adapt-ETdata}
\end{figure*}

\end{document}